\RequirePackage[OT1]{fontenc}
\documentclass[journal]{IEEEtran}

\usepackage{amsmath,amsfonts}
\usepackage{algpseudocode}
\usepackage{algorithm}
\usepackage{array}
\usepackage[caption=false,font=normalsize,labelfont=sf,textfont=sf]{subfig}
\usepackage{subcaption}
\usepackage{textcomp}
\usepackage{stfloats}
\usepackage{url}
\usepackage{verbatim}
\usepackage{graphicx}
\usepackage{cite}
\usepackage{pdfpages}
\usepackage{multirow}
\usepackage{booktabs}
\usepackage[pagebackref,breaklinks,colorlinks]{hyperref}
\usepackage{balance}

\ifCLASSOPTIONcompsoc
  \usepackage[nocompress]{cite}
\else
  \usepackage{cite}
\fi

\newif\ifdiff
\ifdiff
\usepackage[normalem]{ulem}
\def\added#1{{\color{blue}#1}}
\def\removed#1{\textrm{\color{red}\sout{#1}}}
\else
\def\added#1{#1}
\def\removed#1{}
\fi

\begin{document}

\title{\textit{Trident}: Detecting Face Forgeries with \\Domain-adversarial Triplet Learning}

\author{
Mustafa~Hakan~Kara, 
Aysegul~Dundar, 
U\u{g}ur~G\"{u}d\"{u}kbay%,~\IEEEmembership{Senior~Member,~IEEE}%
\IEEEcompsocitemizethanks{
\IEEEcompsocthanksitem M. H. Kara, A. Dundar, and U. G\"{u}d\"{u}kbay are with the Department of Computer Engineering, Bilkent University, Ankara, Turkey.\\
E-mail: hakan.kara@bilkent.edu.tr, \{adundar, gudukbay\}@cs.bilkent.edu.tr
\IEEEcompsocthanksitem Corresponding author: U. G\"{u}d\"{u}kbay.
%\IEEEcompsocthanksitem Manuscript received xxx; revised xxx.
}
}

%\author{%
%\thanks{Manuscript received Month XX, 202X; revised Month XX, 202X.}
%}

\markboth{Kara et al.: Detecting Face Forgeries with Domain-adversarial Triplet Learning}%
{Kara et al.: Detecting Face Forgeries with Domain-adversarial Triplet Learning}

%\IEEEpubid{0000--0000/00\$00.00~\copyright~202X IEEE}

\maketitle

\begin{abstract}
    As face forgeries generated by deep neural networks become increasingly sophisticated, detecting face manipulations in digital media has posed a significant challenge, underscoring the importance of maintaining digital media integrity and combating visual disinformation. Current detection models, predominantly based on supervised training with domain-specific data, often falter against forgeries generated by unencountered techniques. In response to this challenge, we introduce \textit{Trident}, a face forgery detection framework that employs triplet learning with a Siamese network architecture for enhanced adaptability across diverse forgery methods. \textit{Trident} is trained on curated triplets to isolate nuanced differences of forgeries, capturing fine-grained features that distinguish pristine samples from manipulated ones while controlling for other variables. To further enhance generalizability, we incorporate domain-adversarial training with a forgery discriminator. This adversarial component guides our embedding model towards forgery-agnostic representations, improving its robustness to unseen manipulations. In addition, we prevent gradient flow from the classifier head to the embedding model, avoiding overfitting induced by artifacts peculiar to certain forgeries. Comprehensive evaluations across multiple benchmarks and ablation studies demonstrate the effectiveness of our framework. We will release our code in a GitHub repository.

%% UG: avoiding overfitting to artifacts peculiar to certain forgeries. -->
%% UG: avoiding overfitting caused by artifacts peculiar to certain forgeries.
    
\end{abstract}    

\section{Introduction}
Advances in deep-generative networks, particularly Generative Adversarial Networks (GANs)~\cite{goodfellow2014generative} and, more recently, diffusion models~\cite{ho2020denoising}, have led to increasingly sophisticated visual forgeries, posing significant challenges to digital media integrity and public discourse. As manipulated images become prevalent, traditional supervised approaches have shown limitations in:
\begin{itemize}
    \item generalization across diverse manipulation techniques and
    \item detection of previously unseen forgery methods~\cite{wang2020cnn}.
\end{itemize}
To address these challenges, we present \textit{Trident} (\textbf{Tri}plet learning-based \textbf{De}epfake detection \textbf{N}e\textbf{t}work), a triplet learning-based approach designed to generalize across a wide range of forgery techniques, including those not encountered during training. Our method builds upon triplet loss techniques originally introduced for face recognition tasks~\cite{schroff2015} but adapts them to the more complex objective of detecting forged imagery. By carefully curating triplets that preserve the person ID and scene context, \textit{Trident} encourages the model to isolate subtle, forgery-specific features rather than relying on incidental factors like background or identity. This design ensures that the learned embeddings capture the intrinsic differences between genuine and manipulated samples, thereby improving generalization and the detection process's robustness.

\textit{Trident} also leverages domain-adversarial training \cite{ganin2015unsupervised} with a forgery discriminator, which attempts to classify the forgery category given an embedding. This adversarial interplay pushes the embedding generator to produce representations agnostic to specific forgery types, focusing instead on the fundamental artifacts that distinguish real from fake. Such an adversarial mechanism reduces overfitting to known manipulation methods, resulting in discriminative embeddings even when confronted with novel, unseen forgeries. 

In addition, we detach Siamese network embeddings before passing them to the classifier head, ensuring the classifier's gradients do not influence the embedding network during backpropagation. This approach maintains the integrity of the embedding space learned through triplet loss, allowing it to focus more on capturing generalizable forgery artifacts. For our transformer-based variant, we employ BitFit~\cite{ben2022bitfit}, a parameter-efficient fine-tuning approach that prevents catastrophic forgetting while enabling efficient adaptation to the forgery detection task.

Our main contributions to face forgery detection are 
\begin{itemize}
    \item a novel controlled triplet learning formulation that disentangles manipulation cues from identity and scene-specific information by maintaining consistent identity and temporal alignment,
    \item the introduction of an adversarially trained forgery discriminator, guiding the embeddings toward forgery-agnostic representations that remain effective against unseen manipulations,
    \item detaching backbone embeddings from the binary classifier head to further increase generalization, and
    \item efficient bias-only fine-tuning to avoid catastrophic forgetting while adapting to forgery detection with minimal parameter updates.
\end{itemize}
Extensive experiments demonstrate that the \textit{Trident} framework achieves competitive performance on challenging benchmarks while exhibiting more adaptability than traditional supervised approaches. Our findings highlight the potential of combining triplet learning, domain-adversarial training, and parameter-efficient fine-tuning to detect the evolving landscape of face forgeries, contributing to more robust methods for preserving digital media authenticity.
\section{Related Work}

\subsection{Spatial-Domain Methods}

Face-forgery detection methods have evolved significantly to address increasingly sophisticated manipulation techniques~\cite{yan2023deepfakebench}. Early approaches relied on detecting specific artifacts left by generation algorithms. Xception~\cite{chollet2017xception} and EfficientNet~\cite{tan2019efficientnet} emerged as strong baseline models due to their robust performance on benchmark datasets~\cite{rossler2019faceforensics++}. \textit{Face X-Ray}~\cite{li2020face} focused on blending artifacts in face-swapping operations, while Chai \emph{et al.}~\cite{chai2020makes} examined differences between camera-captured and algorithm-manipulated images.

Recent spatial-domain detectors integrate explicit artifact reasoning and finer attention. Li \emph{et al.}~\cite{li2023adal} disentangle forgery artifacts from identity information via adversarial learning, Nguyen \emph{et al.}~\cite{Nguyen_2024_CVPR} employ localized artifact attention to remain quality-agnostic, and Masked-relation learning~\cite{yang2023mrl} models region-to-region relations as a graph, propagating forgery cues globally. Dagar and Vishwakarma~\cite{dagar2024dualbranch} proposed a dual-branch architecture that fuses noise cues with hierarchical ConvNeXt features, attaining superior manipulation-localization accuracy on both shallowfake and deepfake datasets. Yadav and Vishwakarma~\cite{yadav2024awmsa} proposed Face-NeSt, which leverages adaptively weighted multi-scale attentional features and underscores the importance of balanced scale fusion. Long \emph{et al.}~\cite{long2025lgdf} introduced LGDF-Net, a dual-branch fusion network that separately processes local artifact and global texture features through specialized compression and expansion modules.

\subsection{Temporal and Frequency Domain Methods}

Temporal detectors exploit inconsistencies across video frames.  
Yang \emph{et al.}~\cite{yang2019exposing} flagged abnormal head-pose dynamics, while LipForensics~\cite{haliassos2021lips} tracked mouth-movement coherence.  
Gu \emph{et al.}~\cite{gu2021spatiotemporal} proposed Spatio-Temporal Inconsistency Learning, combining spatial cues with a temporal stream; Amerini \emph{et al.}~\cite{Amerini2019opticalflow} leveraged optical-flow CNNs to uncover compression-robust artefacts.  
Choi \emph{et al.}~\cite{choi2024exploiting} detected style-latent incoherence between frames, and \added{Cheng \emph{et al.}~\cite{Cheng2024VoiceFace} introduced a cross-modal strategy that measures voice–face homogeneity, exposing identity mismatches characteristic of face-swap videos.}  
\added{Yu \emph{et al.}~\cite{yu2023amsim} magnify subtle spatio-temporal anomalies via multi-timescale views, while Lu \emph{et al.}~\cite{lu2024ldattention} use long-distance attention to capture global spatial–temporal cues. Xu \emph{et al.}~\cite{Xu_2023_ICCV} introduced TALL, which transforms video clips into pre-defined layouts to preserve spatial-temporal dependencies while being computationally efficient. Li \emph{et al.}~\cite{Li2021Landmarks} leveraged robust facial landmarks with spatial and temporal rotation angles to achieve compression-resistant detection.}

Frequency-domain methods analyze spectral artifacts introduced during synthesis.  
Qian \emph{et al.}~\cite{qian2020thinking} proposed F$^{3}$-Net to mine frequency-aware clues; Li \emph{et al.}~\cite{li2021frequency} learned discriminative features across bands; Liu \emph{et al.}~\cite{liu2021spatial} exploited phase discrepancies.  
Frank \emph{et al.}~\cite{frank2020leveraging} linked GAN up-sampling to high-frequency artefacts.  
\added{Tan \emph{et al.} revisited up-sampling operations by modelling neighbouring-pixel relationships for open-world detection~\cite{Tan_2024_CVPR}; their follow-up FreqNet~\cite{tan2024freqnet} enforces frequency-aware learning to obtain source-agnostic detectors.}

\subsection{Vision Transformer-based Forgery Detection}

Vision Transformers (ViTs) have emerged as powerful architectures for deepfake detection due to their ability to capture long-range dependencies and fine-grained features~\cite{wang2024timelysurveyvisiontransformer}. ISTVT~\cite{zhao2023istvt} combines spatial and temporal attention, while M2TR~\cite{wang2022m2tr} analyzes images at multiple scales through parallel transformer branches. ICT~\cite{dong2022protecting} verifies identity consistency, and UIA-ViT~\cite{zhuang2022uia} models feature distributions without pixel-level masks. Khan \emph{et al.}~\cite{khan2021video} fuse UV texture maps with transformers for pose-invariant detection, and GenConViT~\cite{wodajo2023deepfake} combines generative priors. Forgery-Aware Adaptive Learning ViT (FAL-ViT)~\cite{luo2025falvit} further enhances generalization by adapting a transformer backbone to forgery cues during continual domain shifts. Dagar and Vishwakarma introduced Tex-ViT~\cite{dagar2024texvit}, which augments a ResNet backbone with a texture-aware module and a dual-branch cross-attention Vision Transformer.

\subsection{Generalization-Focused Methods}

\removed{Despite progress in facial forgery detection, generalizing to unseen forgery methods remains challenging.}
\added{A critical challenge in deepfake detection is achieving robustness across unseen manipulations and real-world conditions.} Yu \emph{et al.}~\cite{yu2019attributing} attributed synthetic images to source GANs by learning fingerprints. Ni \emph{et al.}~\cite{ni2022core} enforced representation consistency via \textit{CORE}. Zhao \emph{et al.}~\cite{zhao2021learning} introduced self-consistency learning.  

Huang \emph{et al.}~\cite{huang2023implicit} disentangled identity information, Yan \emph{et al.}~\cite{yan2023ucf} decomposed images into forgery-relevant partitions, and Ojha \emph{et al.}~\cite{ojha2023} used a frozen encoder plus classical classifiers. SBIs~\cite{shiohara2022detecting} employ self-blending, while RECCE~\cite{cao2022end} reconstructs authentic faces. \added{Zhang \emph{et al.}~\cite{zhang2025expansive} mitigate catastrophic forgetting via expansive learning, and cross-modal strategies such as PVASS-MDD~\cite{yu2024pvassmdd} and MCL~\cite{liu2024mcl} leverage alignment and contrastive objectives for improved generalization.}

\subsection{Contrastive and Metric Learning Approaches}

Contrastive and metric learning enhance generalizability by enforcing feature discrimination. Fung \emph{et al.}~\cite{fung2021} proposed DeepfakeUCL, an unsupervised contrastive scheme. Xu \emph{et al.}~\cite{xu2022} developed supervised contrastive learning, while Gu \emph{et al.}~\cite{gu2022} introduced hierarchical video contrast. Liang \emph{et al.}~\cite{liang2023} used depth-guided triplets, and Kumar \emph{et al.}~\cite{kumar2020} applied triplet networks in high-compression scenarios. 

\subsection{Domain-Adversarial Training}

Domain-adversarial training learns domain-invariant representations. Ganin and Lempitsky~\cite{ganin2015unsupervised} introduced the Gradient Reversal Layer for unsupervised adaptation. EANN~\cite{wang2018eann} applied similar ideas to fake news detection. \added{Our work builds on these principles, combining adversarial and triplet objectives to realize forgery-agnostic representations.}
\begin{figure*}[htbp]
    \centering
    \fbox{\includegraphics[width=1.0\textwidth]
    {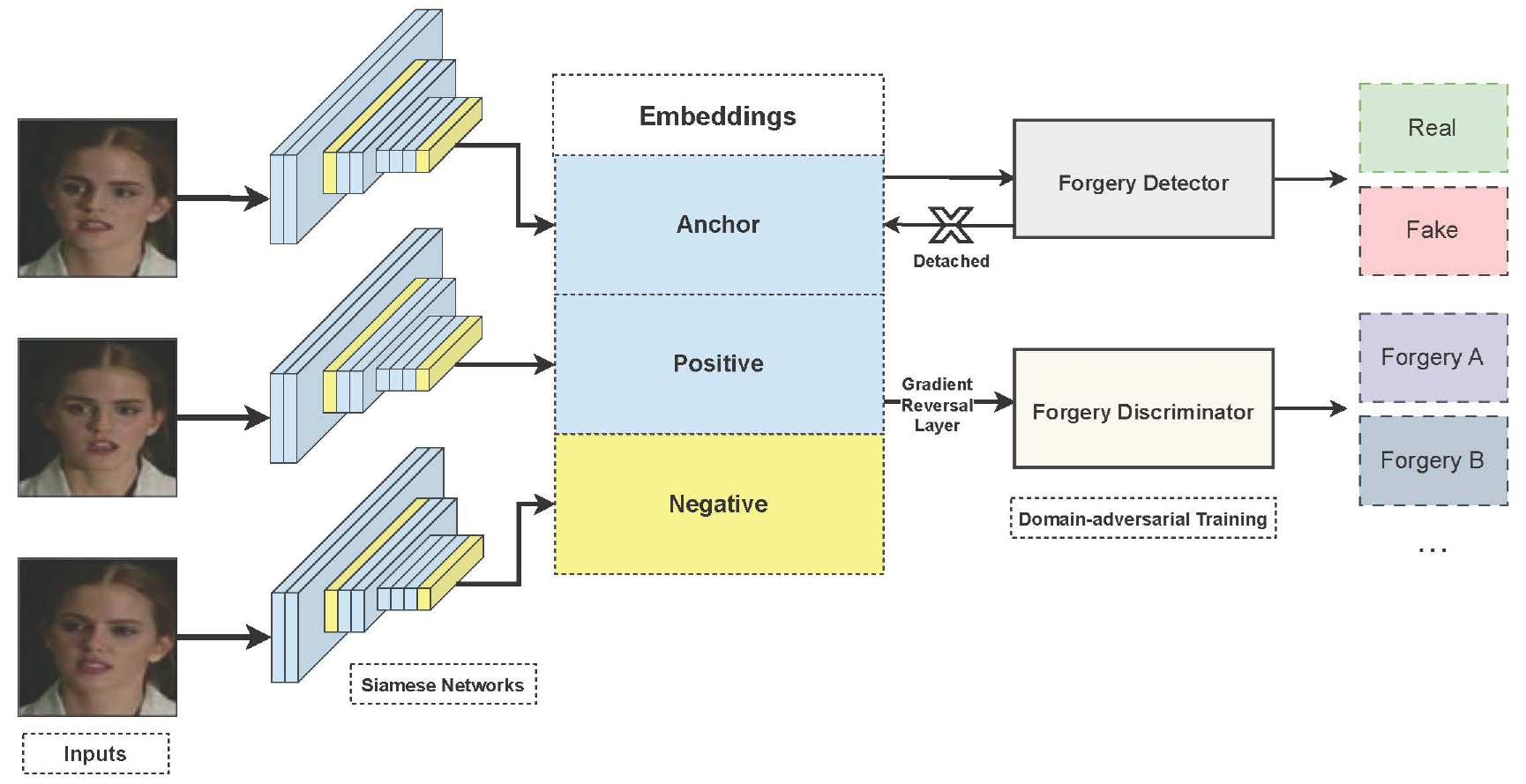}}
    \caption{Overview of our high-level architecture. We leverage a Siamese network with a triplet learning setup to generate embeddings. The forgery detector performs real-fake classification, while the forgery discriminator performs forgery-type classification with a Gradient Reversal Layer to enhance generalization.}
    \label{fig:architecture}
\end{figure*}

\section{Method}
\added{Existing deepfake detection methods often struggle with generalization to unseen forgery techniques due to their reliance on forgery-specific artifacts.} Unlike previous approaches, we apply triplet learning to deepfake detection in a controlled setting. By selecting triplets that maintain a consistent person ID and scene, we isolate forgery-specific features, prompting the model to focus on subtle and discriminative artifacts introduced by the forgery process rather than incidental variations in identity or background.

We also integrate a domain-adversarial training component to drive the network towards learning forgery-agnostic features, employing a forgery discriminator to classify specific manipulation methods. The embedding generator seeks to produce representations indistinguishable to this discriminator, thus pushing the model to rely on universal markers of manipulation rather than forgery-type-specific artifacts. This adversarial interplay leads to more generalized embeddings that better handle emerging and previously unseen forgery techniques.

\subsection{Triplet Learning Framework}
\textit{Trident} employs a triplet learning setting~\cite{TripletEmbedding,schroff2015}, building upon the foundations of contrastive learning and Siamese networks. Contrastive learning, originally proposed for face verification tasks~\cite{chopra2005}, aims to distinguish samples by teaching the model to pull similar items closer and push dissimilar ones apart in the feature space. It employs contrastive loss on pairs of data points, penalizing large distances between similar and small distances between dissimilar pairs~\cite{oh2016deep}.

Siamese networks, introduced by~\cite{bromley1993}, involve feeding pairs of samples into identical neural networks with shared weights. This architecture typically measures the similarity of two inputs regarding learned feature representations. As described in~\cite{schroff2015}, triplet learning extends these concepts by introducing a third element, creating anchor-positive-negative sample triplets. 

This approach aims to bring the anchor and positive samples closer together while pushing the anchor and negative samples farther apart in the feature space. We leverage this triplet learning setting in \textit{Trident}, training a projection head with triplets consisting of two real samples (anchor and positive) and one fake sample (negative). 

We leverage this idea to learn the fine-grained features that distinguish genuine samples from forgeries while controlling for other variables such as scene and person ID. This formulation allows us to capture nuances and forces the model to discern subtle differences and similarities, potentially improving our model's ability to generalize to previously unseen forgeries.

\subsection{Model Architecture}

Our model architecture, shown in Figure~\ref{fig:architecture}, involves a Siamese network structure with a triplet learning setup to distinguish real samples from fake ones. By curating triplets with consistent person ID and scene, we apply triplet learning to deepfake detection in a controlled setting. This controlled triplet learning setting isolates forgery-specific features, allowing the model to focus on finer details and more discriminative features related to the forgery process rather than incidental variations in identity or background. 

\added{The architecture comprises four key components visible in Figure~\ref{fig:architecture}: (1) a \textit{Siamese network} with shared weights that processes triplet inputs and generates embeddings, (2) a \textit{forgery detector} that performs binary real-fake classification on these embeddings, (3) a \textit{forgery discriminator} that attempts to classify forgery types and is connected via a Gradient Reversal Layer (GRL) to promote domain-adversarial training, and (4) a \textit{classifier head detachment mechanism} that isolates the binary classification from embedding generation to preserve generalization capability.} This section provides an in-depth explanation of each component in our architecture.

\subsubsection{Backbone Architecture and Fine-tuning}
We fine-tuned two backbone variants for comprehensive evaluation: EfficientNet-B4~\cite{tan2019efficientnet}, a CNN-based architecture commonly preferred across face forgery detection methods, and CLIP ViT-L/14~\cite{radford2021learning}, OpenAI's transformer-based foundational model pretrained on large-scale visual and textual data.

For the ViT backbone, we employed BitFit~\cite{ben2022bitfit}, a parameter-efficient fine-tuning (PEFT) method that updates only the bias parameters while keeping all weights frozen. BitFit prevents catastrophic forgetting of learned abstract representations from pretraining while enabling efficient adaptation to the forgery detection task. This approach significantly reduces the number of trainable parameters while preserving the model's generalization capability. Table~\ref{tab:backbone_comparison} compares the two backbones we fine-tuned during our experiments.

\begin{table}[htbp]
\centering
\caption{Backbone Comparison}
\label{tab:backbone_comparison}
\begin{tabular}{lcc}
\toprule
\textbf{Backbone} & \textbf{EfficientNet-B4~\cite{tan2019efficientnet}} & \textbf{CLIP ViT-L/14}~\cite{radford2021learning} \\
\midrule
Architecture & CNN & Transformer \\
Fine-tuning & Full & BitFit (bias-only) \\
Embedding Dimension & 1792 & 1024 \\
Total Parameters & 19.3M & 304M \\
Trainable Parameters & 19.3M & 0.3M \\
\bottomrule
\end{tabular}
\end{table}

\subsubsection{Siamese Network and Embedding Generation}

We process each triplet of inputs through three parallel Siamese networks with shared weights, generating embeddings that capture fine-grained, distinguishing features between genuine and forged samples. Our triplet loss structure encourages the model to minimize the distance between the anchor and positive samples while maximizing the distance between the anchor and negative samples in feature space. By designing embeddings like this, we enable the model to learn representations robust to variations such as person ID or scene context, improving its generalization ability.

\subsubsection{Forgery Detector}

We pass these embeddings into the \textit{Forgery Detector}, a binary classifier determining whether a sample is real or fake. By dedicating this head to binary classification, we allow the model to focus on separating real samples from forgeries without directly identifying the forgery type. This separation improves generalization since the Forgery Detector concentrates exclusively on the real versus fake distinction, enabling the model to adapt more effectively to unseen forgery types.

\subsubsection{Domain-adversarial Training Setup}

Our framework employs domain-adversarial training to learn forgery-agnostic representations that generalize across different manipulation techniques. The core principle involves a minimax game between the embedding generator and a forgery discriminator. The embedding generator aims to produce informative features for real-fake classification. However, it is uninformative for forgery-type classification, while the forgery discriminator attempts to classify the specific forgery technique from these embeddings.

This adversarial setup encourages the embedding generator to focus on universal manipulation artifacts rather than technique-specific signatures. The forgery discriminator consists of multiple fully connected layers with an output layer producing logits for each forgery category (Deepfakes, Face2Face, FaceSwap, NeuralTextures, and real). Through this adversarial interaction, the model learns to capture broad real-versus-fake distinctions while avoiding overfitting to particular forgery methods.

\subsubsection{Gradient Reversal Layer}
The Gradient Reversal Layer (GRL)~\cite{ganin2015unsupervised} implements the adversarial training mechanism by reversing gradients during backpropagation. The GRL is defined such that it acts as an identity mapping in the forward pass but reverses and scales the gradient by a factor \(\lambda\) in the backward pass. Let \( x \) be the input embeddings and \(\lambda > 0\):
\begin{equation}
\text{Forward: } G(x) = x
\end{equation}

During the backward pass, the gradient is multiplied by \(-\lambda\):
\begin{equation}
\text{Backward: } \frac{\partial G}{\partial x} = -\lambda I
\end{equation}
\noindent where \( I \) is the identity matrix. This scaling factor \(\lambda\) allows fine-grained control over the adversarial signal.

The GRL is placed between the embedding generator and the forgery discriminator. It acts as a pass-through layer during forward propagation but reverses the gradient from the forgery discriminator's multi-class cross-entropy loss during backpropagation. This reversal encourages the embedding generator to produce feature representations that confuse the forgery discriminator, thereby learning more generalizable, forgery-invariant embeddings. The domain-adversarial training with the GRL aligns the embedding generator's objective with producing domain-invariant features, preventing the model from overfitting to known forgery types.

\subsubsection{Classifier Head Detachment}
We isolate the Forgery Detector (real-fake classification head) from the embedding generation process to further enhance generalization. In practice, we block backpropagation from the forgery detector into the shared embedding layers. By preventing gradient flow from the binary classification task back into the embedding space, we ensure that the embeddings remain focused on forgery-related features rather than overfitting to the particularities of the classification head's decision boundary. This separation helps maintain a more stable and general-purpose embedding space, improving the model's adaptability to previously unseen forgeries.

\subsection{Objective Functions}

The training objective in \textit{Trident} is optimized using a hybrid loss function that combines \textit{Binary Cross-Entropy (BCE) loss}, \textit{Triplet loss}, and \textit{Forgery Discriminator loss}. This composite loss is designed to enhance class discrimination (via BCE), improve feature embedding separation (via Triplet loss), and promote forgery-agnostic embedding generation (via adversarial training through GRL). 

The total loss \( \mathcal{L}_{\text{total}} \) is defined as:
\begin{equation}
\mathcal{L}_{\text{total}} = \mathcal{L}_{\text{BCE}} + \alpha \cdot \mathcal{L}_{\text{triplet}} + \beta \cdot \mathcal{L}_{\text{forgery}}
\end{equation}

\noindent where \( \alpha \) and \( \beta \) are weighting factors that balance the contributions of the individual loss components.

\subsubsection{Binary Cross-Entropy Loss (BCE)}
The Binary Cross-Entropy loss \( \mathcal{L}_{\text{BCE}} \) is defined as:
\begin{equation}
\mathcal{L}_{\text{BCE}} = -\frac{1}{N} \sum_{i=1}^{N} \left( y_i \cdot \log(\hat{y}_i) + (1 - y_i) \cdot \log(1 - \hat{y}_i) \right)
\end{equation}

\noindent where \( N \) is the number of samples, \( y_i \) is the ground truth label, and \( \hat{y}_i \) is the predicted probability of the sample being a forgery. This loss encourages accurate classification by minimizing the error between the predicted and actual labels.

\subsubsection{Triplet Loss}
The Triplet loss \( \mathcal{L}_{\text{triplet}} \) is formulated as:
\begin{equation}
\mathcal{L}_{\text{triplet}}{=}\frac{1}{N} \sum_{i=1}^{N} \left[ \|f(x_i^a){-}f(x_i^p)\|_2^2{-}\|f(x_i^a){-}f(x_i^n)\|_2^2{-}m \right]_+
\end{equation}

\noindent where \( x_i^a \), \( x_i^p \), and \( x_i^n \) represent the anchor, positive, and negative samples, respectively, and \( m \) is the margin that enforces a minimum separation between positive and negative pairs. This loss function brings the anchor and positive embeddings closer together while pushing the anchor and negative embeddings farther apart, thereby enhancing feature distinctiveness.

\subsubsection{Forgery Discriminator Loss}
The forgery discriminator loss \( \mathcal{L}_{\text{forgery}} \) employs multi-class cross-entropy to classify forgery types:
\begin{equation}
\mathcal{L}_{\text{forgery}} = -\frac{1}{N} \sum_{i=1}^{N} \sum_{j=1}^{K} y_{i,j} \log(\hat{p}_{i,j})
\end{equation}

\noindent where \( y_{i,j} \) is the one-hot encoded ground truth for the \( j \)-th forgery category, \( \hat{p}_{i,j} \) is the predicted probability of the \( i \)-th embedding belonging to the \( j \)-th forgery category, and \( K \) is the number of forgery categories. During backpropagation, the GRL reverses the gradients from this loss, encouraging the embedding generator to produce forgery-agnostic representations.

\subsection{Triplet Formation}

\added{Our controlled triplet formation strategy represents a key methodological contribution that distinguishes our approach from standard triplet learning-based methods.} We constructed triplets comprising anchor, positive, and negative samples to train our model effectively. This approach encourages the network to learn embeddings where genuine and manipulated images of the same individual are mapped to distinct regions in the feature space. We formed the triplets as follows:

\begin{itemize}
\item \textit{Anchor}: An image (real or fake) of a specific individual.
\item \textit{Positive}: Same individual, same authenticity as anchor, different timestamp.
\item \textit{Negative}: Same individual, same timestamp as anchor, opposite authenticity.
\end{itemize}

We partitioned the real and fake samples into two halves for each identity. Denote these partitions as $R_1$, $R_2$ (real samples) and $F_1$, $F_2$ (fake samples). We ensured that the anchor and negative samples were temporally aligned. We then systematically generated all possible triplet configurations for each identity, alternating which partition serves as anchor, positive, and negative samples, as illustrated in Algorithm~\ref{alg:triplet_generation}.

\renewcommand{\algorithmicrequire}{\textbf{Input:}}
\renewcommand{\algorithmicensure}{\textbf{Output:}}

\begin{algorithm}
\caption{Triplet Generation}
\label{alg:triplet_generation}
\begin{algorithmic}[noend] % Use [noend] to avoid "end" statements
\Function{FormTriplets}{RealSamples,FakeSamples}
    \State Divide RealSamples into two parts: $R_1$ and $R_2$
    \State Divide FakeSamples into two parts: $F_1$ and $F_2$
    \State Initialize empty collections for Triplets and Labels
    \For{\textbf{each} identity}
        \State Add $(R_1, R_2, F_1)$ to Triplets with label $(0, 0, 1)$
        \State Add $(F_1, F_2, R_1)$ to Triplets with label $(1, 1, 0)$
        \State Add $(R_2, R_1, F_2)$ to Triplets with label $(0, 0, 1)$
        \State Add $(F_2, F_1, R_2)$ to Triplets with label $(1, 1, 0)$
    \EndFor
    \State \Return Triplets, Labels
\EndFunction
\end{algorithmic}
\end{algorithm}

This triplet formation constitutes the main novelty of our approach, allowing us to control for identity, scene, and temporal factors. We aim to disentangle manipulation cues from identity or scene-specific information by maintaining consistent identity and temporal alignment while varying authenticity.
\section{Experiments}
\label{sec:experiments}
We evaluated the proposed method across three experimental scenarios using the FaceForensics++ (FF++) dataset~\cite{rossler2019faceforensics++} and conducted cross-dataset evaluation on Google's Deepfake Detection (DFD) dataset~\cite{dfd_dataset}. This section details the experimental setups, implementation specifics, and the configurations employed for the ablation studies.

\subsection{Area Under the Curve}
\label{subsec:auc}
To evaluate the performance of our method, we employ the \textit{Area Under the Receiver Operating Characteristic Curve (AUC)}, a widely used metric in deepfake detection literature. The AUC provides a threshold-independent measure of a classifier's ability to distinguish between real and fake samples.
\subsubsection{AUC and the ROC Curve}
The AUC is computed as the integral of the Receiver Operating Characteristic (ROC) curve by computing the area under the ROC curve, which plots the \textit{True Positive Rate (TPR)} against the \textit{False Positive Rate (FPR)} at various classification thresholds.
\subsubsection{Computation of AUC}
The ROC curve is integrated using the trapezoidal rule to compute AUC numerically. Given a set of \(n\) thresholds, the AUC is computed using the trapezoidal rule:
\begin{equation}
\text{AUC}{=}\sum_{k=1}^{n-1} \frac{(\textit{FPR}_{k+1}{-}\textit{FPR}_k) \cdot (\textit{TPR}_k{+}\textit{TPR}_{k+1})}{2}
\label{eq:auc_trapezoidal}
\end{equation}
\noindent where \(\textit{FPR}_k\) and \(\textit{TPR}_k\) represent the values of the False Positive Rate and True Positive Rate at the \(k\)$^\textit{th}$ threshold. AUC is particularly suitable for deepfake detection due to its robustness to class imbalance and threshold independence.

\subsection{Datasets}
To evaluate the generalization capability of our approach, we employed several benchmark datasets with diverse forgery techniques and quality levels:

\textit{FaceForensics++ (FF++)}~\cite{rossler2019faceforensics++} contains 1,000 authentic videos alongside their corresponding manipulated versions generated through four distinct facial manipulation techniques: Deepfakes~\cite{deepfakes2018}, Face2Face~\cite{thies2016face2face}, FaceSwap~\cite{faceswap2018}, and NeuralTextures~\cite{thies2019deferred}. The dataset provides videos at multiple compression levels, enabling comprehensive evaluation across diverse quality degradation scenarios.

\textit{Celebrities DeepFake (Celeb-DF)}~\cite{li2020celebdf} includes two versions: \textit{Celeb-DF (v1)} consists of 408 real videos and 795 corresponding DeepFake videos, and \textit{Celeb-DF (v2)} significantly expands upon this, comprising 590 real videos and 5,639 corresponding DeepFake videos with minimal visual artifacts. Version 2 is particularly challenging for detection due to its realistic visual quality, closely matching real-world DeepFake videos circulated online.

\textit{DeepFake Detection (DFD)}~\cite{dfd_dataset} released by Google contains thousands of Deepfake videos featuring consenting actors in controlled environments. It encompasses various facial expressions, head poses, and lighting conditions.

\textit{DeepFake Detection Challenge (DFDC)}~\cite{dolhansky2020deepfake} released by Facebook contains over 100,000 videos created using various deepfake techniques. It features diverse subjects across different ages, ethnicities, and lighting conditions. We also use its preview version \textit{DeepFake Detection Challenge Preview (DFDCP)}~\cite{dolhansky2019deepfake}, which contains a smaller subset of videos but maintains similar diversity, making both datasets particularly challenging for generalization.

\textit{University at Albany DeepFake Videos (UADFV)}~\cite{yang2019exposing} includes authentic and deepfake manipulated videos specifically curated to expose inconsistencies in head poses—a common artifact in early deepfake synthesis methods.

Table~\ref{tab:dataset_breakdown} provides a concise summary of these datasets, including the number of pristine (original) and manipulated (fake) samples. 

\begin{table}[!t]
    \footnotesize
    \renewcommand{\arraystretch}{1.2}
    \setlength{\tabcolsep}{3pt}
    \caption{Breakdown of Benchmarks}
    \label{tab:dataset_breakdown}
    \centering
    \begin{tabular}{lccc}
    \hline
    \textbf{Dataset} & \textbf{Pristine} & \textbf{Manipulated} & \textbf{Total} \\
    \hline
    Celeb-DF (v1)~\cite{li2020celebdf}       & 408   & 795    & 1,203  \\
    Celeb-DF (v2)~\cite{li2020celebdf}       & 590   & 5,639  & 6,229  \\
    FaceForensics++~\cite{rossler2019faceforensics++} & 1,000 & 4,000  & 5,000  \\
    DFD~\cite{dfd_dataset}            & 363   & 3,000  & 3,363  \\
    DFDC~\cite{dolhansky2020deepfake}           & 19,154 & 86,654 & 105,808 \\
    DFDCP~\cite{dolhansky2019deepfake}         & 1,131  & 4,119  & 5,250  \\
    UADFV~\cite{yang2019exposing}          & 49    & 49     & 98     \\
    \hline
    \end{tabular}
\end{table}
    
\subsection{Evaluation Scenarios}
We designed three experimental scenarios to evaluate the effectiveness of the proposed method and analyze the contributions of its components:

\begin{itemize}
\item \textit{Scenario 1: Forgery-Specific Training}. Models were trained on only two forgery types (\textit{Deepfakes} and \textit{NeuralTextures}) from the FF++ dataset. The trained models were then tested on all forgery types in FF++ to assess their ability to generalize to unseen manipulation techniques. This scenario examined the impact of the Gradient Reversal Layer (GRL) and Triplet Learning.

\item \textit{Scenario 2: Full Dataset Training}. Models were trained and tested on all forgery types in the FF++ dataset. This scenario measured the overall performance of the proposed method when exposed to a diverse set of forgeries during training.

\item \textit{Scenario 3: Cross-Dataset Evaluation}. Models trained on the FF++ (HQ) dataset were evaluated on the DFD dataset. This scenario tested the models' generalization to a completely different dataset with variations in data distribution and manipulation techniques.
\end{itemize}

\subsection{Implementation Details}

\subsubsection{Network Architecture}

We employed both EfficientNet-B4~\cite{tan2019efficientnet} (CNN) and MARLIN~\cite{cai2023marlin} (ViT) as backbone architectures for feature extraction to explore complementary representation capabilities.

We also designed a \texttt{NetworkTree} structure to implement the proposed architecture, enabling flexible configuration and easy insertion or modification of components. The overall network architecture forms a tree with the root being the embedding model and two child modules:

\begin{itemize}
\item \textit{Forgery Detector}: A binary classifier that predicts whether an input image is real or fake, utilizing the embedding from the feature extractor.
\item \textit{Forgery Discriminator}: An auxiliary classifier that predicts the forgery type, used with the GRL to encourage the embedding model to learn features invariant to specific forgery techniques.
\end{itemize}

\subsubsection{Training Configuration}
\begin{table}[htbp]
\caption{Hyperparameter Comparison}
\centering
\renewcommand{\arraystretch}{1.2}
\begin{tabular}{lcc}
\hline
\textbf{Backbone} & \textbf{EfficientNet-B4~\cite{tan2019efficientnet}} & \textbf{CLIP ViT-L/14}~\cite{radford2021learning} \\ \hline
Learning rate & $0.0001$ & $0.00002$ \\
Batch size & $4$ & $8$ \\
Triplet Loss margin (\( m \)) & $1.0$ & $1.0$ \\
GRL lambda ($\lambda$) & $1.0$ & $1.0$ \\
Forgery loss weight & $1.0$ & $0.5$ \\
Number of epochs & $30$ & $7$ \\
Optimizer & Adam~\cite{kingma2015adam} & Adam~\cite{kingma2015adam} \\
\hline
\end{tabular}
\label{tab:hyperparameters}
\end{table}

We employed different hyperparameter configurations for our CNN and ViT variants, as shown in Table~\ref{tab:hyperparameters}. The ViT variant requires a lower learning rate (0.00002 vs. 0.0001) and fewer epochs (7 vs. 30) than the CNN variant due to parameter-efficient fine-tuning. Both variants share identical triplet margin and GRL parameters.
\section{Results}
\label{sec:results}

This section presents the experimental results of our framework. The performance is evaluated quantitatively through ablation studies combined with intra-dataset and cross-dataset evaluations. Qualitative analysis is provided through t-SNE visualization of the learned feature embeddings.

\subsection{Quantitative Results}
\begin{table*}[htbp]
   \caption{Frame-level \textbf{AUC (\(\uparrow\))} Scores of Face Forgery Detectors. The Best Results are Indicated in Bold.}
   \centering
   \renewcommand{\arraystretch}{1.2}
   \small
   \resizebox{\textwidth}{!}{
   \begin{tabular}{l|l|l|l|ccccc|cccccc}
   \toprule
   \textbf{Type} & \textbf{Detector} & \textbf{Backbone} & \textbf{Venue} & 
   \multicolumn{5}{c|}{\textbf{Intra-Dataset Evaluation on FF++ (HQ)}} & 
   \multicolumn{6}{c}{\textbf{Cross-Dataset Evaluation}} \\
   \cmidrule(lr){5-9} \cmidrule(lr){10-15}
    & & & & \textbf{FF++ (HQ)} & \textbf{FF-DF} & \textbf{FF-F2F} & \textbf{FF-FS} & \textbf{FF-NT} & 
   \textbf{Celeb-v1} & \textbf{Celeb-v2} & \textbf{DFD} & \textbf{DFDCP} & \textbf{DFDC} & \textbf{UADFV} \\
   \midrule
   
   \multirow{5}{*}{\textbf{Naive}}
   & MesoNet~\cite{afchar2018mesonet} 
      & Designed CNN 
      & WIFS-2018 
      & 0.6077 
      & 0.6771 
      & 0.6170 
      & 0.5946 
      & 0.5701 
      & 0.7358 
      & 0.6091 
      & 0.5481 
      & 0.5994
      & 0.5560
      & 0.7150 \\
   
   & MesoInception~\cite{afchar2018mesonet} 
      & Designed CNN 
      & WIFS-2018 
      & 0.7583 
      & 0.8542 
      & 0.8087 
      & 0.7421 
      & 0.6517 
      & 0.7366 
      & 0.6966 
      & 0.6069 
      & 0.7561
      & 0.6226
      & 0.9049 \\
   
   & CNN-Aug~\cite{wang2020cnn} 
      & ResNet~\cite{he2016deep} 
      & CVPR-2020 
      & 0.8493 
      & 0.9048 
      & 0.8788 
      & 0.9026 
      & 0.7313 
      & 0.7420 
      & 0.7027 
      & 0.6464 
      & 0.6170
      & 0.6361
      & 0.8739 \\
   
   & Xception~\cite{rossler2019faceforensics++} 
      & Xception~\cite{chollet2017xception} 
      & ICCV-2019 
      & 0.9637 
      & 0.9799 
      & 0.9785 
      & 0.9833 
      & 0.9385 
      & 0.7794 
      & 0.7365 
      & 0.8163 
      & 0.7374
      & 0.7077
      & 0.9379 \\
   
   & EfficientNet-B4~\cite{tan2019efficientnet} 
      & EfficientNet~\cite{tan2019efficientnet} 
      & ICML-2019 
      & 0.9567 
      & 0.9757 
      & 0.9758 
      & 0.9797 
      & 0.9308 
      & 0.7909 
      & 0.7487 
      & 0.8148 
      & 0.7283
      & 0.6955
      & 0.9472 \\
   
   \midrule
   
   \multirow{3}{*}{\textbf{Frequency}}
   & F3Net~\cite{qian2020thinking} 
      & Xception~\cite{chollet2017xception} 
      & ECCV-2020 
      & 0.9635 
      & 0.9793 
      & 0.9796 
      & 0.9844 
      & 0.9354 
      & 0.7769 
      & 0.7352 
      & 0.7975 
      & 0.7354
      & 0.7021
      & 0.9347 \\
   
   & SPSL~\cite{liu2021spatial} 
      & Xception~\cite{chollet2017xception} 
      & CVPR-2021 
      & 0.9610 
      & 0.9781 
      & 0.9754 
      & 0.9829 
      & 0.9299 
      & 0.8150
      & 0.7650
      & 0.8122 
      & 0.7408
      & 0.7040
      & 0.9424 \\
   
   & SRM~\cite{luo2021generalizing} 
      & Xception~\cite{chollet2017xception} 
      & CVPR-2021 
      & 0.9576 
      & 0.9733 
      & 0.9696 
      & 0.9740 
      & 0.9295 
      & 0.7926 
      & 0.7552 
      & 0.8120 
      & 0.7408
      & 0.6995
      & 0.9427 \\
   
   \midrule
   
   \multirow{12}{*}{\textbf{Spatial}}
   & Capsule~\cite{nguyen2019capsule} 
      & CapsuleNet~\cite{sabour2017dynamic} 
      & ICASSP-2019 
      & 0.8421 
      & 0.8669 
      & 0.8634 
      & 0.8734 
      & 0.7804 
      & 0.7909 
      & 0.7472 
      & 0.6841 
      & 0.6568
      & 0.6465
      & 0.9078 \\
   
   & DSP-FWA~\cite{li2018exposing} 
      & Xception~\cite{chollet2017xception} 
      & CVPRW-2019 
      & 0.8765 
      & 0.9210 
      & 0.9000 
      & 0.8843 
      & 0.8120 
      & 0.7897 
      & 0.6680 
      & 0.7403 
      & 0.6375
      & 0.6132
      & 0.8539 \\
   
   & Face X-ray~\cite{li2020face} 
      & HRNet~\cite{wang2020deep} 
      & CVPR-2020 
      & 0.9592 
      & 0.9794 
      & \textbf{0.9872} 
      & 0.9871
      & 0.9290 
      & 0.7093
      & 0.6786
      & 0.7655 
      & 0.6942
      & 0.6326
      & 0.8989 \\
   
   & FFD~\cite{dang2020detection} 
      & Xception~\cite{chollet2017xception} 
      & CVPR-2020 
      & 0.9624 
      & 0.9803 
      & 0.9784 
      & 0.9853 
      & 0.9306 
      & 0.7840
      & 0.7435
      & 0.8024 
      & 0.7426
      & 0.7029
      & 0.9450 \\
   
   & CORE~\cite{ni2022core} 
      & Xception~\cite{chollet2017xception} 
      & CVPRW-2022 
      & 0.9638 
      & 0.9787 
      & 0.9803 
      & 0.9823 
      & 0.9339 
      & 0.7798
      & 0.7428
      & 0.8018 
      & 0.7341
      & 0.7049
      & 0.9412 \\
   
   & RECCE~\cite{cao2022end} 
      & Custom 
      & CVPR-2022 
      & 0.9621 
      & 0.9797 
      & 0.9779 
      & 0.9785 
      & 0.9357 
      & 0.7677
      & 0.7319
      & 0.8119 
      & 0.7419
      & 0.7133
      & 0.9446 \\
   
   & UCF~\cite{yan2023ucf} 
      & Xception~\cite{chollet2017xception} 
      & ICCV-2023 
      & 0.9705 
      & 0.9883
      & 0.9840 
      & \textbf{0.9896} 
      & 0.9441 
      & 0.7793
      & 0.7527
      & 0.8074 
      & 0.7594
      & 0.7191
      & 0.9528 \\
   
   & LSDA~\cite{Yan_2024_CVPR}$^*$ 
      & EfficientNet~\cite{tan2019efficientnet} 
      & CVPR-2024 
      & 0.9549 
      & 0.9738 
      & 0.9772 
      & 0.9691 
      & 0.8994 
      & 0.8689
      & 0.8014
      & 0.8296 
      & 0.7810
      & 0.7330
      & 0.9009 \\

    & Effort~\cite{yan2025orthogonal}$^*$ 
      & CLIP ViT~\cite{radford2021learning} 
      & ICML-2025 
      & 0.9083 
      & \textbf{0.9907} 
      & 0.9124 
      & 0.9843 
      & 0.7459 
      & 0.9041
      & 0.8497
      & 0.8991 
      & 0.8133
      & 0.8211
      & 0.9659 \\

     % --- Our method (\textit{Trident} CNN) ---
   & \textit{Trident} (CNN) 
     & EfficientNet~\cite{tan2019efficientnet} 
     & Ours 
     & \textbf{0.9793} 
     & 0.9779 & 0.9843 & 0.9831 
     & \textbf{0.9590} 
     & 0.7609 
     & 0.7543 & 0.8740 
     & \textbf{0.8403}
     & 0.6976
     & 0.9450 \\

     % --- Our method (\textit{Trident} CLIP ViT) ---
   & \textit{Trident} (ViT) 
     & CLIP ViT~\cite{radford2021learning} 
     & Ours 
     & 0.9266 
     & 0.9899 & 0.9441 & 0.9803 
     & 0.7920 
     & \textbf{0.9115} 
     & \textbf{0.8606} & \textbf{0.9204} 
     & 0.8263
     & \textbf{0.8476}
     & \textbf{0.9666} \\
      
   \bottomrule
   \end{tabular}
   }
   \label{tab:combined_evaluation}
   \vspace{0.5em}
   \footnotesize{$^*$Reproduced with the official implementations due to unavailability of weights on \text{DeepfakeBench}.}
\end{table*}

Table~\ref{tab:combined_evaluation} presents face forgery detection performance under standardized \text{DeepfakeBench}~\cite{yan2023deepfakebench} evaluations. All the methods are trained on FF++ (HQ) dataset. Cross-dataset results reveal significant performance degradation across methods, highlighting generalization challenges. To ensure a fair comparison, all metrics are obtained through \text{DeepfakeBench}'s controlled evaluation setting.

\subsubsection{Ablation Studies}

Ablation studies were conducted to assess the impact of key components in the \textit{Trident} framework with the CNN backbone, specifically Triplet Learning (TL), the Gradient Reversal Layer (GRL), the Adversarial Loss (Adv), and the Detached Classification Head (DH). Experiments were performed under two settings:

\begin{enumerate}
    \item training and testing on all FF++ (HQ) dataset forgery types and
    \item training on only \text{Deepfakes} and \text{NeuralTextures} manipulations, then testing on all forgery types in the FF++ (HQ) dataset.
\end{enumerate}

%The results are presented in Tables~\ref{tab:ablation_all} and~\ref{tab:ablation_dfnt}, respectively.

\paragraph{Ablation Study on All Forgery Types}

Table~\ref{tab:ablation_all} summarizes the model's training and testing results on all forgery types in the FF++ (HQ) dataset. The baseline model (\textbf{B}) employs the EfficientNet-B4 backbone with a binary classification head. The variants include the addition of Triplet Learning (\textbf{TL}) and the incorporation of the Gradient Reversal Layer (\textbf{TL{+}GRL}).

\begin{table}[htbp]
\caption{Ablation Study on FF++ (HQ) Dataset (All Forgery Types)}
\centering
\renewcommand{\arraystretch}{1.2}
\begin{tabular}{lcc}
\hline
\textbf{Method} & \textbf{AUC (\(\uparrow\))} & \textbf{LogLoss (\(\downarrow\))} \\\hline
\textbf{Baseline} & 0.9497 & 0.2915 \\
\textbf{TL} & 0.9613 & 0.2611 \\
\textbf{TL{+}GRL} & 0.9669 & 0.2946 \\
\textbf{TL{+}GRL{+}DH} & \textbf{0.9793} & \textbf{0.2456} \\
\hline
\end{tabular}
\label{tab:ablation_all}
\end{table}

\paragraph{Ablation Study on \text{Deepfakes} and \text{NeuralTextures}}

In this setting, the models are trained only on \text{Deepfakes} and \text{NeuralTextures} manipulations and tested on all forgery types in the FF++ (HQ) dataset. Table~\ref{tab:ablation_dfnt} presents the results. The methods include the baseline (\textbf{B}), the addition of Triplet Learning (\textbf{TL}), the incorporation of the Gradient Reversal Layer (\textbf{TL{+}GRL}), the use of Adversarial Loss (\textbf{TL{+}Adv}), and the inclusion of the Detached Classification Head (\textbf{TL{+}GRL{+}DH}).

\begin{table}[htbp]
\caption{Ablation Study on FF++ (HQ) Dataset (\text{Deepfakes} and \text{NeuralTextures})}
\centering
\renewcommand{\arraystretch}{1.2}
\begin{tabular}{lcc}
\hline
\textbf{Method} & \textbf{AUC (\(\uparrow\))} & \textbf{LogLoss (\(\downarrow\))} \\
\hline
\textbf{Baseline} & 0.7363 & 0.9512 \\
\textbf{TL} & 0.7506 & 0.8645 \\
\textbf{TL{+}Adv} & 0.9595 & 0.3270 \\
\textbf{TL{+}GRL} & \textbf{0.9652} & 0.2662 \\
\textbf{TL{+}GRL{+}DH} & 0.9646 & \textbf{0.2454} \\
\hline
\end{tabular}
\label{tab:ablation_dfnt}
\end{table}

\paragraph{Cross-Dataset Ablation on DFD (HQ)}
To further assess the impact of our proposed components on generalization ability, we conducted an ablation study on the DFD (HQ) dataset with models trained on FF++ (HQ). Table~\ref{tab:cross_ablation_dfd} presents the results. The \textit{LogLoss} refers to the Binary Cross-Entropy (BCE) loss used for the classification task in both tables.

\begin{table}[htbp]
\caption{Cross-Dataset Ablation on DFD (HQ) Dataset (Training on FF++ (HQ))}
\centering
\renewcommand{\arraystretch}{1.2}
\begin{tabular}{lcc}
\hline
\textbf{Method} & \textbf{AUC (\(\uparrow\))} & \textbf{LogLoss (\(\downarrow\))} \\
\hline
\textbf{Baseline} & 0.7817 & 0.6897 \\
\textbf{TL} & 0.8438 & \textbf{0.5522} \\
\textbf{TL{+}GRL} & 0.8668 & 0.6076 \\
\textbf{TL{+}GRL{+}DH} & \textbf{0.8740} & 0.5568 \\
\hline
\end{tabular}
\label{tab:cross_ablation_dfd}
\end{table}

\subsection{Qualitative Results}
\begin{figure}[htbp]
\centering
\subfloat[Baseline (B)]{\includegraphics[width=0.50\columnwidth]{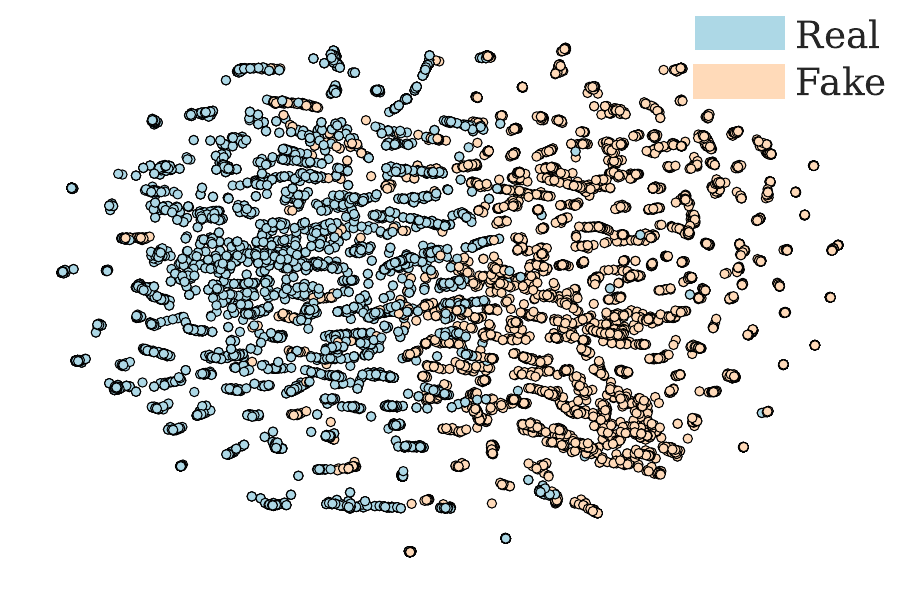}}\hfill
\subfloat[B+GRL+DH]{\includegraphics[width=0.50\columnwidth]{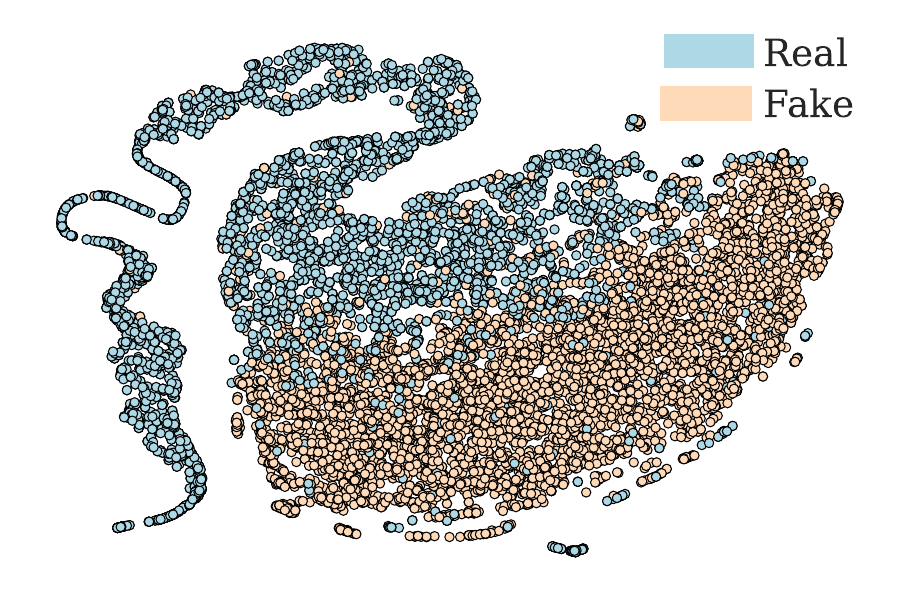}}
\caption{t-SNE embeddings for models without TL. GRL+DH strengthens real/fake separation but does not yield five distinct forgery clusters.}
\label{fig:tsne_notl}
\vspace{-0.5em}
\end{figure}

\begin{figure}[htbp]
\centering
\subfloat[TL]{\includegraphics[width=0.33\columnwidth]{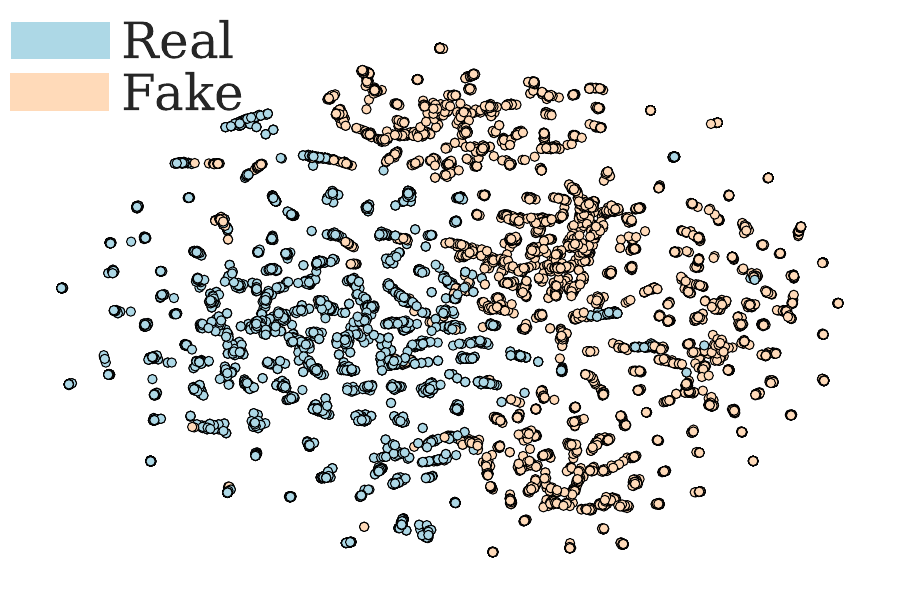}}\hfill
\subfloat[TL+GRL]{\includegraphics[width=0.33\columnwidth]{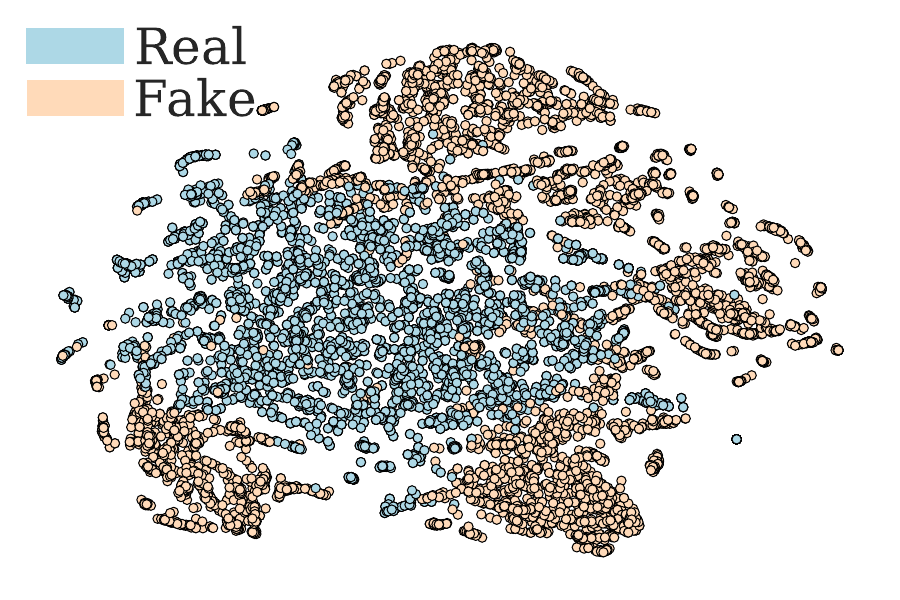}}\hfill
\subfloat[TL+GRL+DH]{\includegraphics[width=0.33\columnwidth]{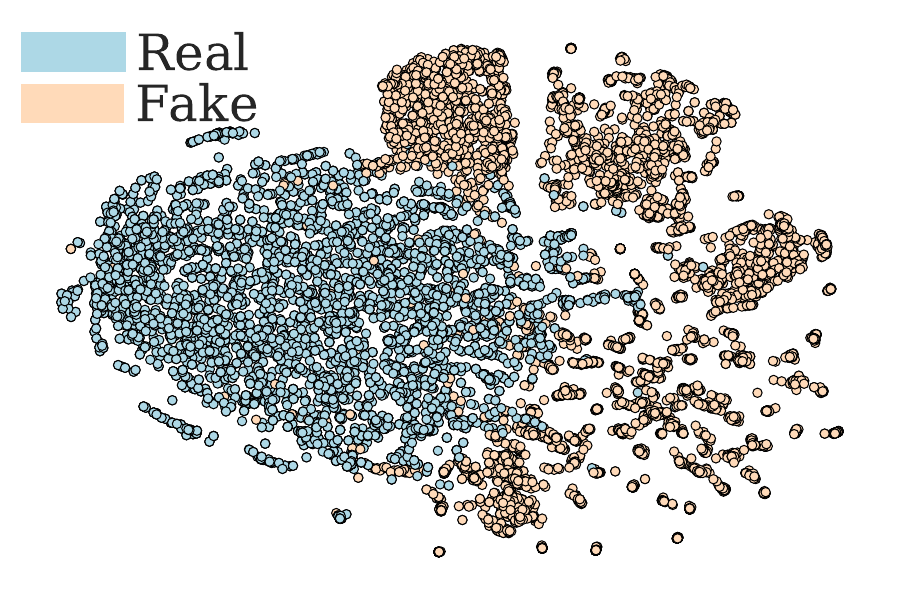}}
\caption{t-SNE embeddings for TL-based models. TL promotes forgery-type clustering; GRL and DH enhance real/fake separation.}
\label{fig:tsne_tl}
\vspace{-1em}
\end{figure}

To qualitatively assess the discriminative capability of the learned embeddings, we employed t-distributed Stochastic Neighbor Embedding (t-SNE)~\cite{maaten2008visualizing} to project high-dimensional features into two-dimensional space. We present two sets of visualizations: baseline models in Figure~\ref{fig:tsne_notl} and Triplet Learning-based models in Figure~\ref{fig:tsne_tl}. Real and fake samples are represented in blue and orange, respectively. All visualizations are produced from the FF++ (HQ) validation split.

Figure~\ref{fig:tsne_notl} illustrates embeddings for models \emph{without} Triplet Learning (TL). In Fig.~\ref{fig:tsne_notl}(a) (Baseline, B), real (blue) and fake (orange) samples overlap considerably, indicating weak discriminative power. Adding Gradient Reversal Layer (GRL) and Detached Head (DH) components (B+GRL+DH, Fig.~\ref{fig:tsne_notl}(b)) strengthens real/fake separation but still does not yield five clear clusters for the FF++ forgeries.

In contrast, Fig.~\ref{fig:tsne_tl} shows \emph{TL-based} models. As seen in Fig.~\ref{fig:tsne_tl}(a) (TL), incorporating TL creates distinct clusters for each forgery type, though some overlap remains. Employing GRL and DH components alongside TL (TL+GRL+DH, Fig.~\ref{fig:tsne_tl}(b)) further separates real from fake while forming five distinct clusters corresponding to the five forgery types present in the FF++ dataset.

\section{Discussion}
\label{sec:discussion}

The experimental results presented in Section~\ref{sec:results} provide several important insights into face forgery detection:

\paragraph{Performance Comparison}
The evaluation results in Table~\ref{tab:combined_evaluation} demonstrate that \textit{Trident} (CNN) achieves 0.9793 AUC on FF++ (HQ), surpassing the previous SOTA approach, UCF (0.9705). In cross-dataset evaluation, our CNN variant achieves notable improvements on DFD (0.8740 vs.\ Xception's 0.8163) and DFDCP (0.8403 vs.\ UCF's 0.7594). The ViT variant outperforms previous methods on multiple benchmarks, demonstrating the best performance on Celeb-v1 (0.9115 vs.\ Effort's 0.9041), Celeb-v2 (0.8606 vs.\ Effort's 0.8497), DFD (0.9204 vs.\ Effort's 0.8991), DFDC (0.8476 vs.\ Effort's 0.8211), and UADFV (0.9666 vs.\ Effort's 0.9659). This consistent performance improvement across different datasets highlights the mitigation effect of our framework against the generalization challenge in face forgery detection.

\paragraph{Component Contributions}
The ablation studies quantify each component's impact. From Tables~\ref{tab:ablation_all} and~\ref{tab:ablation_dfnt}, Triplet Learning provides a foundational improvement, raising the baseline AUC from 0.9497 to 0.9613 with all forgery types and from 0.7363 to 0.7506 with limited forgery types. The most significant gain occurs when combining TL with adversarial techniques (GRL/Adv): when trained only on \text{Deepfakes} and \text{NeuralTextures} but tested on all forgery types, TL+GRL achieves 0.9652 AUC—a 22.89\% increase over the baseline's 0.7363. This demonstrates the framework's ability to learn forgery-agnostic features. The cross-dataset results in Table~\ref{tab:cross_ablation_dfd} reinforce this finding, showing that TL+GRL+DH achieves 0.8740 AUC on DFD, surpassing the baseline by 11.8\%.

\paragraph{Feature Space Organization}
The t-SNE visualizations in Figures~\ref{fig:tsne_notl} and~\ref{fig:tsne_tl} reveal the models' discriminative capabilities. Models without Triplet Learning show considerable overlap between real and fake samples, even with GRL and DH components. TL-based models produce more distinct clustering by forgery type, with the full \textit{Trident} configuration (TL+GRL+DH) achieving both clear real/fake separation and distinct clustering for different forgery techniques. 

\paragraph{CNN vs. ViT Backbone}
The results reveal distinct performance patterns between backbone architectures. The CNN-based \textit{Trident} achieves superior intra-dataset performance while demonstrating strong results on DFD and DFDCP benchmarks. In contrast, the ViT-based variant excels across most cross-dataset scenarios, showing consistent improvement on the majority of evaluation benchmarks. This performance divergence suggests that each architecture captures different aspects of forgery artifacts. The CNN's hierarchical feature extraction appears effective for controlled intra-dataset scenarios, while the ViT architecture's self-attention mechanism and BitFit fine-tuning strategy demonstrate superior transferability across diverse cross-domain scenarios, likely due to better preservation of pretrained representations.

\paragraph{Impact of BitFit Fine-tuning}
The ViT variant's strong cross-dataset performance highlights the effectiveness of BitFit fine-tuning in preserving learned representations while adapting to forgery detection. By updating only bias parameters, BitFit prevents catastrophic forgetting of abstract visual features learned during CLIP pretraining, enabling better generalization to unseen datasets and forgery types. This parameter-efficient approach proves particularly valuable in scenarios with limited training data or when adapting to new domains.

\paragraph{Generalization Challenges}
Despite significant improvements achieved, cross-dataset performance reveals ongoing generalization challenges in deepfake detection. The performance variation across different test sets suggests that current methods, including ours, remain sensitive to dataset-specific characteristics such as compression artifacts, capture conditions, and forgery implementation details. However, our \textit{Trident} framework demonstrates more consistent cross-domain performance compared to existing approaches, indicating that controlled triplet learning combined with domain-adversarial training provides a promising direction for addressing these fundamental challenges in forgery detection generalization.
\section{Conclusion}
We introduce \textit{Trident}, a forgery detection framework that integrates triplet learning and domain-adversarial training to address the growing sophistication and diversity of face forgery methods. Our approach isolates subtle forgery-specific features by leveraging triplet learning with carefully curated samples that share identity and scene but differ in authenticity. This controlled representation learning strategy produces robust embeddings that retain discriminative power across forgery techniques. Simultaneously, our domain-adversarial training scheme with a dedicated forgery discriminator produces forgery-agnostic embeddings that capture generalizable manipulation markers.

We implement two backbone variants: EfficientNet and CLIP ViT with BitFit fine-tuning. The parameter-efficient BitFit approach prevents catastrophic forgetting while enabling effective adaptation to forgery detection tasks with minimal computational overhead.

Experiments demonstrate that CNN-based \textit{Trident} achieves state-of-the-art FF++ (HQ) performance, surpassing existing methods on DeepfakeBench. Our ViT variant demonstrates superior cross-dataset generalization, outperforming existing methods on multiple benchmarks, including Celeb-v1, Celeb-v2, DFD, DFDC, and UADFV. The ablation studies confirm that combining triplet learning, gradient reversal layers, and classifier head detachment significantly enhances detection performance. The t-SNE visualizations further validate our approach by demonstrating a clear separation between real and fake samples with distinct clustering by forgery type.

Our results reveal complementary strengths between backbone architectures: CNN-based \textit{Trident} excels in intra-dataset scenarios with strong performance on specific benchmarks, while ViT-based \textit{Trident} demonstrates superior transferability across diverse cross-domain scenarios. The effectiveness of BitFit fine-tuning in preserving pretrained representations while adapting to forgery detection highlights the value of parameter-efficient approaches in this domain.

While \textit{Trident} shows notable improvements in cross-dataset performance, the remaining gap between intra-dataset and cross-dataset results indicates persistent generalization challenges in face forgery detection. Future research directions include incorporating temporal cues, multi-modal inputs, and advanced metric learning approaches. Investigating adaptive triplet sampling strategies or dynamic margin selection could refine the embedding space and address the remaining generalization challenges.
\balance


%\input{author_bio}
\end{document}